\documentclass[3p,10pt,a4paper,twoside,fleqn,sort&compress]{elsarticle}

\makeatletter
\def\ps@pprintTitle{%
  \let\@oddhead\@empty
  \let\@evenhead\@empty
  \let\@oddfoot\@empty
  \let\@evenfoot\@oddfoot
} \makeatother

\usepackage{graphicx}
\usepackage{verbatim}
\usepackage{amsmath}
\usepackage[varg]{pxfonts}
\usepackage{eepic}
\usepackage{amssymb}

\newtheorem{theorem}{Theorem}[section]

\newtheorem{example}[theorem]{Example}

\newcommand{\mb}[1]{\mathbb{#1}}
\newcommand{\bs}[1]{\boldsymbol{#1}}
\newcommand{\mc}[1]{\mathcal{#1}}
\newcommand{\ra}{\rightarrow}

\newcommand{\pd}{\partial}

\newcommand{\bc}[2]{\binom{\ #1 \ }{\ #2 \ }}

\newcommand{\sR}{\mb R}
\newcommand{\sK}{\mb K}

\newcommand{\sN}{\mb N}
\newcommand{\deriv}[3]{#1 \overset{#2}{\Rightarrow} #3}

\newcommand{\mN}{\mc N}
\newcommand{\mR}{\mc R}

\newcommand{\tM}{\text M}

\newcommand{\mgfc}[3]{M_{#1,#2 | #3}(\vt)}

\newcommand{\mom}[2]{\mu_{p, \vX_{#2}}^{(#1)}}
\newcommand{\moms}[2]{\mu_{p, \vX_{#2}}^{(#1)}}
\newcommand{\momc}[3]{\mu_{p, \vX_{#2} | \bs{#3}}^{(#1)}}
\newcommand{\momcs}[3]{\mu_{p, \vX_{#2} | \bs{#3}}^{(#1)}}
\newcommand{\momvec}[2]{\mathbf \mu_{#1}^{(#2)}}

\renewcommand\leq{\leqslant}

\newcommand{\mia}{\bs \alpha}
\newcommand{\mib}{\bs \beta}
\newcommand{\mig}{\bs \gamma}
\newcommand{\minu}{\bs \nu}
\newcommand{\Ainf}{\sN_0^d}

\newcommand{\sA}{\mc A}

\newcommand{\vt}{\bs t}
\newcommand{\vX}{\bs X}

\newcommand{\egf}[2]{\sum_{\bs #1 \in \Ainf} \frac{#2^{(#1)}}{#1 !} \cdot \vt^{(\bs #1)}}

\newcommand{\bsel}[3]{\big(\ #1^{(\bs #2)}\ \big)_{#2 \in #3} }

\newcommand{\mgf}[2]{M_{#1,#2}(\vt)}

\newcommand{\Anu}{A_{\bs \nu}}
\newcommand{\hbegf}[3]{\mc B^{(#1)} \big\{\ \egf{#2}{#3} \ \big\}}
\newcommand{\cA}{\mc A}
\newcommand{\ot}{\otimes}
\newcommand{\osum}{\bigoplus}
\newcommand{\oprod}{\bigotimes}

\begin{document}

\begin{frontmatter}

\title{\Large\bf Cross-moments computation for stochastic context-free grammars\tnoteref{t1}}
\tnotetext[t1]{Research supported by Ministry  of Science and Technological Development, Republic of Serbia, Grants No. 174013 and 174026}

\author[misanu,fsmun]{Velimir M. Ili\'c\corref{cor}}
\ead{velimir.ilic@gmail.com}

\author[fsmun]{Miroslav D. \'Ciri\'c}
\ead{miroslav.ciric@pmf.edu.rs}

\author[fosun]{Miomir S. Stankovi\'c}
\ead{miomir.stankovic@gmail.com}

\cortext[cor]{Corresponding author. Tel.: +38118224492; fax: +38118533014.}
\address[misanu]{Mathematical Institute of the Serbian Academy of Sciences and Arts, Kneza Mihaila 36, 11000 Beograd, Serbia}
\address[fsmun]{University of Ni\v s, Faculty of Sciences and Mathematics, Vi\v segradska 33, 18000 Ni\v s, Serbia}
\address[fosun]{University of Ni\v s, Faculty of Occupational Safety, \v Carnojevi\'ca 10a, 18000 Ni\v s, Serbia}

\begin{abstract}
In this paper we consider the problem of efficient computation of
cross-moments of a vector random variable represented by a
stochastic context-free grammar.~Two types of cross-moments are
discussed.~The~sample space for the first~one is the set of all
derivations of the context-free grammar, and the sample space for
the second one is the set of all derivations which generate a
string belonging to the language of the grammar.~In the past, this
problem was widely studied, but mainly for the cross-moments of
scalar variables~and up to the second order.~This paper presents
new algorithms for computing the cross-moments of an arbitrary
order, and the previously developed ones are derived as special
cases.
\end{abstract}


\end{frontmatter}


\section{Introduction}

The cross-moments of random variables modeled with stochastic
context-free grammars (\textit{SCFG}) are~important quantities in
the \textit{SCFG} modeling \cite{Hutchins_72}. They are defined as
expected value of the product of integer powers of the entries of
random vector variable, which can represent string or derivation
length, the number of rule occurrences in a derivation or
uncertainty associated with the occurring rule.~The expec\-tation
can be taken either with respect to the sample space of all
\textit{SCFG} derivations or with respect to the sample space of
all derivations which generate a string belonging to the language
of the grammar.~Throughout this paper, the name
\textit{cross-moments} is usually used in the former case, while
in the latter case we talk~about \textit{conditional
cross-moments}.

The computation of cross-moments may become demanding if the
sample space is large.~In the past,~this problem was widely
studied, but mainly for the cross-moments of scalar variables
(called simply~\textit{moments}) and up to the second order.~The
first order moments computation, such as expected length of
derivations and expected string length, are given in
\cite{Wetherell_80}. The computation of \textit{SCFG} entropy is
considered in \cite{Nederhof_Satta_08c}. The procedure for
computing the moments of string and derivation length is given in
\cite{Hutchins_72}, where the explicit~formulas for the moments up
to the second order are derived.~First order conditional
cross-moments~are~considered in \cite{Hwa_00}, where the algorithm
for conditional \textit{SCFG} entropy is derived.~A more general
algorithm for computing the conditional cross-moments of a vector
variable of the second order is derived in \cite{Li_Eisner_09}.

In this paper we give the recursive formulas for computing the
cross-moments and the conditional~cross-moments of an arbitrary
order, for a vector variable which factorizes according to a
certain rule which is satisfied in the case of string or
derivation length, the number of rule occurrences in derivation or
uncertainty associated with the occurring rule.
The formulas are derived by the differentiation of the recursive
equations for the moment generating function \cite{Lund_04}, which
are obtained from the algorithms for computing the partition
function of a \textit{SCFG} \cite{Nederhof_Satta_08a} for the
cross-moments and with the inside algorithm \cite{Lari_Young_90},
\cite{Goodman_99} for the conditional cross-moments.

The paper is organized as follows.~Section 2 introduces
multi-index notation, which is used throughout the paper, and
reviews some preliminary notions about generalized Leibniz's
formula, basic algebraic~structures, and context free grammars.~In
Section 3 we give the formal definition of \textit{SCFG}
cross-moments and moment-generating function.~The recursive
equations for cross-moments are given in Section 4 for the case
when the sample space is the set of all derivations, while in
Section 5 we consider  the set of all derivations which generate a
string belonging to the language of the  grammar as the sample
space.

\section{Preliminaries}

This section gives some basic definitions and theorems which are
used in the paper. We review the multiindex formulation of the
\textit{Generalized Leibniz's} formula \cite{Raymond_91}, and
basic notions from the theory of weighted context free grammars,
according to \cite{Nederhof_Satta_08a} and
\cite{Nederhof_Satta_08b}.

\subsection{Multiindexes, Multinomial theorem and Generalized Leibniz's formula}

\textbf{Multi-indexes.} A multi-index is defined as a tuple of
nonnegative integers $\bs \alpha = (\alpha_1, \dots , \alpha_d)
\in \sN_0^d$.
We define its dimension as $\dim (\bs \alpha) = d$ and its length
as the sum $| \bs \alpha | = \alpha_1+\alpha_2+ \cdots +
\alpha_d$. The multi-index factorial is $\bs \alpha ! = \alpha_1!
\cdots \alpha_d !$. The zero multi-index is $\bs 0 =(0,\dots,0)$.

If $\bs \beta = (\beta_1, \dots , \beta_d) \in \sN_0^d$, we write
$\bs \beta < \bs \alpha$ if $\beta_i < \alpha_i$ for $i = 1,
\dots, d$. We write $\bs \beta \leq \bs \alpha$ provided $\beta_i
\leq \alpha_i$ for $i = 1, \dots, d$. 
The sum and difference of $\bs \alpha$ and $\bs \beta$, where
$\mib \leq \mia$, are defined to be $\bs \alpha \pm \bs \beta =
(\alpha_1 \pm \beta_1, \dots ,\alpha_d \pm \beta_d)$.

If $\mib_1, \dots, \mib_N$ are multi-indexes and $\bs \beta_1 +
\cdots + \bs \beta_N = \bs \alpha$, we define the multinomial
coefficients to be
\begin{equation*}
\label{prel: mind: multinom}
\bc{\bs \alpha}{\bs \beta_1, \dots , \bs \beta_N} =%
\frac{\bs \alpha !}{\bs \beta_1 ! \cdots  \bs \beta_N !}.
\end{equation*}
For a vector $\bs z= (z_1, \dots , z_d) \in \sR^d$ and a
multi-index $\bs \beta = (\beta_1, \dots, \beta_d) \in \sN_0^d$,
the multi-index power is defined to be
\begin{equation*}
\label{prel: mind: multipower} \bs z^{\bs \beta}=z_1^{\beta_1}
\cdots  z_d^{\beta_d}.
\end{equation*}

\textbf{Multinomial theorem and Generalized Leibniz's formula.}
With these settings, the multinomial theorem \cite{Protter_98} can
be expressed as
\begin{equation*}
\label{prel: mind: multinomial theorem}
\Big(\sum_{i=1}^N \bs z_i \Big)^{\bs \alpha}=%
\sum_{\bs \beta_1 + \cdots + \bs \beta_N = \bs \alpha} \bc{\bs \alpha}{\bs \beta_1, \dots , \bs \beta_N} \prod_{i=1}^N \bs z_i^{\bs \beta_i},%
\end{equation*}
for a vector $\bs z= (z_1, \dots , z_d) \in \sR^d$ and $\bs \alpha
= (\alpha_1, \dots, \alpha_d) \in \sN_0^d$.

Let $\mia=(\alpha_1, \dots, \alpha_d)$ and let $C_{\bs \alpha}$
denote the set of all functions $u:\sR^d \ra \sR$ having $\mia$-th
partial derivative. For a function $u:\sR^d \ra \sR$, we define
the partial derivative of an order $\mia$ as
\begin{equation*}
\label{prel: glr: D_nu}%
\mc D^{(\mia)} \big\{ u(t_1, \dots, t_d) \big\}= \frac{\pd^{|\bs
\alpha|} u(t_1, \dots, t_d)}{\pd^{\alpha_1}t_1 \dots
\pd^{\alpha_d}t_d}\bigg|_{t=0}.
\end{equation*}

Note that $\mc D^{(\bs 0)} \big\{ u(\bs t) \big\} = u(\bs
t)$.~According to the \textit{generalized Leibniz's formula}
\cite{Raymond_91}, the following equality holds
\begin{equation}
\label{prel: glr: D_FG} \mc D^{(\mia)} \big\{ FG  \big\} =
\sum_{\bs 0 \leq \bs \beta \leq \bs \alpha} \bc{\bs \alpha}{\bs
\beta} \ \mc D^{(\mib)}\big\{ F \big\} \cdot \mc D^{(\mia-\mib)}
\big\{ G \big\},
\end{equation}
for all $F,G \in C_{\bs \alpha}$. The derivative of the product of
more than two functions can be found according to
\cite{Thaheema_Laradjia_03}
\begin{equation}
\label{prel: glr: D_prod_F}
\mc D^{(\mia)} \Big\{ \prod_{i=1} ^m F_i \Big\} =%
\sum_{\bs \beta_1 + \cdot + \bs \beta_m = \bs \nu} \bc{\bs \alpha}{\bs \beta_1, \dots , \bs \beta_m}%
\prod_{i=1} ^m  \mc D^{(\mib_i)} \big\{ F_i \big\},
\end{equation}
for all $F_i \in C_{\bs \nu};\ i=1, \dots , m$.

\textbf{Tuples of elements indexed with multi-indexes}.
The set of all multi-indexes lower than or equal to $\bs \nu$ is
denoted with $\sA_{\bs \nu}$,
\begin{equation*}
\sA_{\bs \nu}=\big\{ \bs \alpha \in \sN_0^{\dim(\bs \nu)} \ | \
\bs \alpha \leq \bs \nu  \big\},
\end{equation*}
and $|\sA_{\bs \nu}|$ denotes its cardinality.

For $\bs \alpha = (\alpha_1, \dots , \alpha_d)$ and $\bs \beta =
(\beta_1, \dots , \beta_d)$, we define the lexicographic order
relation $\prec$, so that $\bs \alpha \prec \bs \beta$ if
\begin{equation*}
\alpha_1 = \beta_1, \dots , \alpha_{n}  = \beta_{n} \text{ and }
\alpha_{n+1} < \beta_{n+1}.
\end{equation*}

Let $\bs \nu = (\nu_1, \dots, \nu_d) \in \sN_0^d$ be a multi-index
and $\bs \alpha_1, \dots, \bs \alpha_{|\sA_{\bs\nu}|}$ be
multi-indexes from $\sA_{\bs\nu}$ such that $\bs 0 =\bs \alpha_1
\prec \bs \alpha_2 \prec \cdots \prec \bs \alpha_{|\sA_{\bs \nu}|}
= \bs \nu$.
%
Let $\bs z = (z_1, \dots, z_{|\sA_{\minu}|}) \in
\sR^{|\sA_{\minu}|}$ and let $z: \sA_{\bs\nu} \ra \sR$ be a
function which to each $\bs \alpha_i$ from $\sA_{\bs \nu}$
associates a real number $z^{(\mia_i)}$, such that
$z^{(\mia_i)}=z_i$. We use the following notation for vector $\bs
z$
\begin{equation*}
\bs z = \big(\ z^{(\bs \alpha)}\ \big)_{\bs \alpha \in \sA_{\bs
\nu}} = (z^{(\bs \alpha_1)}, \dots , z^{(\bs \alpha_{|\sA_{\bs
\nu}|})}).
\end{equation*}

\subsection{Semirings}
A \textit{monoid} is a triple $(\mb K,\oplus, 0)$ where $\oplus$
is an associative binary operation on the set $\mb K$ and $0$ is
the identity element for $\oplus$, i.e. $a \otimes 0 = 0 \oplus a
= a$, for all $a \in \mb K$. A monoid is commutative if the
operation $\oplus$ is commutative.

\begin{example}\rm
Let $\Sigma$ be a non-empty set. The \emph{free monoid} $\mathbf \Sigma^*=(\Sigma, \cdot, \epsilon)$ over $\Sigma$ is a monoid,
where the carrier set $\Sigma^* = \{\,a_1\dots a_n \,|\, n \in \sN_0,\, a_i \in \Sigma\,(1\leq i \leq n) \}$ is the set of all \emph{strings}
over $\Sigma$ and $\epsilon$ is the (unique) empty string~of length zero. The operation $\cdot$ denotes the composition (concatenation) of strings defined by
$\bs u_1 \cdot \bs u_2 = \bs u_1 \bs u_2$ for all $\bs u_1, \bs u_2 \in \Sigma^*$.
\end{example}

A \textit{semiring} is a tuple $(\mb K,\oplus,\otimes, 0, 1)$ such that
\begin{enumerate}
\item $(\mb K,\oplus,0)$ is a commutative monoid with $0$ as the identity element for $\oplus$,
\item $(\mb K,\otimes,1)$ is a monoid with $1$ as the identity element for $\otimes$,
\item $\otimes$ distributes over $\oplus$, i.e.
$(a \oplus b) \otimes c = (a \otimes c) \oplus ( b \otimes c)$ and
$c \otimes (a \oplus b) = (c \otimes a) \oplus ( c \otimes b)$,
for all $a, b, c$ in $\mb K$,\item $0$ is an annihilator for
$\otimes$, i.e. $a \otimes 0 = 0 \otimes a = 0$, for every $a$ in
$\mb K$.
\end{enumerate}
A semiring is commutative if the operation  $\otimes$ is commutative.~The
operations $\oplus$ and $\otimes$ are called the addition and the
multiplication in $\mb K$. For a topology $\tau$ we define the topological semiring as a pair
$\big(\mb K, \tau \big)$.

\textbf{Semiring of multivariate power series}.~A $d$-dimensional
formal power series is defined as a map $s: \sN_0^d \ra \sR$ and
can be denoted as an infinite tuple
\begin{equation*}
\bs s = \big(\ s^{(\bs \alpha)}\ \big)_{\mia \in \sN_0^d}.
\end{equation*}
Let $\sR\big[ \sN_0^d \big]$ denote the set of all $d$-dimensional
formal power series:
\begin{equation*}
\sR\big[ \sN_0^d \big]= \Big\{ s: \sN_0^d \ra \sR \Big\}.
\end{equation*}
In order to create a semiring structure, we write elements of
$\sR\big[ \sN_0^d \big]$ as
\begin{equation*}
\bs s(\vt)= \sum_{\bs \alpha \in \Ainf}%
s^{(\bs \alpha)} \bs t^{\bs \alpha},
\end{equation*}
where $\vt \in \sR^d$, 
and use the usual rules for power series,
\begin{align}
\label{mgfmp: s_1+s_2}
(\bs s_1+ \bs s_2)(\vt)&=%
\sum_{\bs \alpha \in \Ainf}%
(s_1^{(\bs \alpha)}+ s_2^{(\bs \alpha)}) \bs t^{\bs \alpha},\\
\label{mgfmp: s_1*s_2}
(\bs s_1\cdot \bs s_2)(\vt)&=%
\sum_{\mia \in \Ainf} \Big(\sum_{\mib + \mig = \mia}%
s_1^{(\mib)}s_2^{(\mig)}\Big) \vt^{\mia}.
\end{align}
A commutative semiring of multivariate power series can now be
defined as the tuple $\big( \sR\big[ \sN_0^d \big], +, \cdot, 0, 1
\big)$, where the addition and multilication are defined with
(\ref{mgfmp: s_1+s_2})-(\ref{mgfmp: s_1*s_2}) and the identities
$0$ and $1$ are from $\sR$.

\subsection{Weighted and stochastic context-free grammars}

By a \textit{weighted context-free grammar (WCFG)} over a
commutative semiring $\big(\mb K, +, \cdot, 1, 0 \big)$ we mean a
tuple $G = \big( \Sigma, \mN, S, \mR , w \big)$, where

\begin{itemize}

\item $\Sigma = \big\{ w_1, \dots, w_{|\Sigma|}\big\}$ is a finite set of
\emph{terminals},

\item $\mN = \big\{ A_1, \dots, A_{|\mN|}\big\}$ is a finite set of
\emph{nonterminals} disjoint with $\Sigma$,

\item $S \in \mN$ is called the \emph{start symbol} (throughout the paper it is usually assumed that $S=A_1$),

\item $\mR \subseteq \mN \times (\Sigma \cup \mN)^{*} $ is a finite set
of rules. A rule $(A, \alpha) \in \mR$ is commonly written as
$A \ra \alpha$, where the nonterminal $A$ is called the \emph{premise}. The set of all rules $A_i \ra B_{i,j}$,
$B_{i,j}\in (\mN \cup \Sigma)^*$ will be denoted by $\mR_i$.

\item $w: R \ra \sK$ is the function called \textit{weight}.

\end{itemize}

The \textit{left-most rewriting relation} $\Rightarrow$ associated
with $G$ is defined as the set of triples $\big( \alpha, \pi,
\beta \big) \in (\Sigma \cup \mN)^* \times \mR \times (\Sigma \cup
\mN)^*$, for which there is a terminal string $\bs u \in \Sigma^*$
and a nonterminal string $\delta \in (\Sigma \cup \mN)^*$, along
with a nonterminal $A \in \mN$ and a string $\gamma \in (\Sigma
\cup \mN)^*$ such that $\alpha = \bs u A \delta$, $\beta = \bs u
\gamma \delta$, and $\pi = A \ra \gamma$ is a rule~from~$\mR$. The
left-most relation triple $\big( \alpha, \pi, \beta \big)$ will be
denoted by $\deriv{\alpha}{\pi}{\beta}$. The \emph{left-most
derivation} (hereinafter the \emph{derivation}) in this grammar is
a string $\pi_1, \dots, \pi_n \in \mR^*$ for which there are
grammar symbols $ \alpha, \beta\in \Sigma \cup \mN$ such that we
can derive $\beta$ from $\alpha$ by applying the rewriting rules
$\pi_1, \dots, \pi_n$
$\deriv{\alpha}{\pi_1}{\deriv{\cdots}{\pi_n}{\beta}}$.  The weight
function is extended to derivations such that $w\ (\pi_1 \cdots
\pi_N) = w(\pi_1) \cdots w(\pi_N)$, for all $\pi_1 \cdots \pi_N
\in R^*$. A nonterminal $A$ is \emph{productive} if there exists a
derivation $\pi_1\cdots\pi_k$ such that
$\deriv{A}{\pi_1}{\deriv{\cdots}{\pi_k}{\bs u}}, \bs u \in
\Sigma^*$.~A nonterminal~$A$ is \emph{accessible} from a
nonterminal $B$ if there exist derivations $\pi_1\cdots\pi_k$ such
that $\deriv{B}{\pi_1}{\deriv{\cdots}{\pi_k}{\eta A \xi}}$ where
$\eta,\xi \in (\Sigma \cup \mN)^*$ (if $A$ is accessible from $S$,
then it is simply accessible). A nonterminal $A$ is \emph{useful}
if it is accessible and productive (otherwise, it is
\emph{useless}).

A weighted context-free grammar $G =\big(\Sigma, \mN, A_1, \mR , p
\big)$ over the probability semiring $\big( \mb R_+, +, \cdot, 0,1
\big)$ is~called~a \emph{stochastic context-free grammar}
(\textit{SCFG}) if the weight $p$ maps all rules to the real unit
interval $[0,1]$. A \emph{SCFG} is \emph{reduced} if $p(A \ra
\gamma)>0$ for all $A\ra \gamma \in \mR$ and each nonterminal $A$,
and all nonterminals are useful. In this paper we consider only
the reduced \emph{SCFGs}. In addition, we assume that the
\emph{SCFG} is \emph{proper}, which means that the weight function
$p$ gives us a probability distribution over the rules that we can
apply, i.e.~$\sum_{j=1}^{|\mR_i|} p\big( A_i \ra
B_{i,j})=1$~for~all $1 \leq i \leq |\mN|$.

For a stochastic context-free grammar $G =\big(\Sigma, \mN, A_1,
\mR , p \big)$ we define the subgrammar $G_i= \big( \Sigma,
\mN_i', A_i, \mR_i' , p_i' \big)$ with the start symbol $A_i$,
where $\mN_i'$ is the set which consists of $A_i$ and nonterminals
accessible from $A_i$ and $\mR_i' \subseteq \mR$ is the set of
rules in which only nonterminals from $\mN_i'$ appear as premises
and $p_i'(\pi)=p(\pi)$ for each $\pi \in \mR_i'$. In the following
text, for the notational convenience, we will assume $p \equiv
p_i'$ when there is no danger of confusion. Note that if $G_i$ is
reduced, then $G_i$ also has this property.

\section{Moment-generating function of \textit{SCFG}}
\label{mgf}

Let $G = \big(\Sigma, \mN, A_1, \mR , p \big)$ be a stochastic
context-free grammar, $\Omega$  the set of all derivations in $G$,
and $\Omega_i$ the set of all derivations starting at $A_i \in
\mN$. The grammar $G$ is \emph{consistent} if
\begin{equation*}
\sum_{\bs \pi \in \Omega_i} p (\bs \pi) = 1,
\end{equation*}
for $1 \leq i \leq |\mN|$. Booth and Thompson \cite{Booth_Thompson_73} gave the consistency
condition for the start symbol $S=A_1$ by the following theorem.

\begin{theorem}
\label{cm: theo: consistency} A reduced stochastic context-free
grammar $G$ is consistent if $\rho(\tM)<1$, where $\rho(\tM)$ is
the absolute value of the largest eigenvalue of the expectation
matrix $\tM=[\tM_{i,n}], 1 \leq i,n \leq |\mN|$ defined by
\begin{equation}
\label{cm: M_in} \tM_{i,n}= \sum_{j=1}^{|\mR_i|} p\big(A_i \ra
B_{i,j}\big) r_n(i,j),
\end{equation}
where $r_n(i,j)$ denotes the number of times  the nonterminal
$A_n$ appears on the right-hand side of the rule $\pi= A_i \ra
B_{i,j}$.
\end{theorem}

Note that the expectation matrices $\tM^{(i)}$ of all subgrammars
$G_i$ are the principal submatrices of $\tM$, and according to
\cite{Horn_Johnson_85} (Corollary 8.1.20),
$\rho(\tM^{(i)})\leq\rho(\tM)$ and $G_i$ are also consistent, i.e.

\begin{equation}
\label{cm: consistency_i}
\sum_{\bs \pi \in \Omega_i} p (\bs \pi)=1.
\end{equation}

Let $G =\big(\Sigma, \mN, A_1, \mR , p \big)$ be SCFG, let $G_i=
\big( \Sigma, \mN_i', A_i, \mR_i' , p \big)$ be $i$-th subgrammar,
$i=1,\cdots,|\mc N|$, and let $\bs X_i = \big[ X_{i,1}(\bs \pi),
\dots , X_{i,D}(\bs \pi) \big]^T$ be random variables distributed
according to the $p_i'$ (recall that $p_i'(\bs \pi)=p(\bs \pi)$).

The \emph{$i$-th cross-moment of an order $\bs \nu = (\nu_1, \dots
, \nu_D)$} is defined with
\begin{equation}
\label{mgf: m_nu_def} \mom{\nu}{i}=
\sum_{\bs \pi \in \Omega_i} p(\bs \pi) \cdot %
X_{i,1}(\bs \pi)^{\nu_1} \cdots X_{i,D}(\bs \pi)^{\nu_d} = %
\sum_{\bs \pi \in \Omega_i} p(\bs \pi)\ \bs X_i(\bs \pi)^{\bs
\nu}.
\end{equation}

The direct computation of (\ref{mgf: m_nu_def}) by enumerating all
derivations is inefficient, since it requires the $\mc
O(|\Omega|)$ operations, and it even becomes infeasible when
$\Omega$ is an infinite set. On the other hand, if we can derive
the expressions for efficient computation of the moment-generating
function (\ref{mgf: M_X_def}), the moment can be retrieved by
differentiation.

In this paper we consider the random vectors $\bs X$ which can be
represented as the sum of random vectors $\bs Y : \mc R \ra \mb
R$:
\begin{equation}
\label{mgf: X_sum_Y}%
\bs X(\pi_1 \cdots  \pi_N) =%
\bs Y\big( \pi_1 \big) + \cdots + \bs Y\big( \pi_N \big) ,
\end{equation}
for all $\pi_1 \cdots  \pi_N \in \Omega$. This assumption may seem
too restrictive, but it holds in some important cases: (1) If $\bs
X(\bs \pi)$ represents derivation length, then $\bs Y(\pi_i)=1$;
(2) if $\bs X(\bs \pi)$ is the derived string length, then $\bs
Y(\pi_i)$ equals the number of terminals on the right-hand side of
$\pi_i$; (3) if $\bs X(\bs \pi)$ represents the self-information
of derivation $\bs \pi$ \cite{Hwa_00}, then $\bs Y(\pi_i)=-\log
p(\pi_i)$.

Following the Proposition 6 from \cite{Chi_99}, it can be shown
that the cross-moments are bounded if the factorization (\ref{mgf:
X_sum_Y}) holds and, for all $\bs t=(t_1, \dots, t_D)$; $|t_i|<1$,
we have
\begin{equation}
\sum_{\bs \pi \in \Omega_i} p(\bs \pi)\ \big(\bs t^T\bs X_i(\bs
\pi)\big)^{\bs \nu}<\sum_{\bs \pi \in \Omega_i} p(\bs \pi)\ \bs
X_i(\bs \pi)^{\bs \nu}< C <\infty \quad \Rightarrow \quad%
\sum_{k=0}^\infty \frac{1}{k!} \ \sum_{\bs \pi \in \Omega_i}
p(\bs\pi) \big(\bs t^T\bs X_i(\bs \pi)\big)^k =%
\sum_{\bs \pi \in \Omega_i} p(\bs \pi) e^{\bs t^T \vX_i(\bs \pi)}.
\end{equation}
Accordingly, we can define the \textit{$i$-th moment-generating
function (MGF)}, as the function $M_{p,\bs X_i}: \mb R^D \ra \sR$,
where
\begin{equation}
\label{mgf: M_X_def}
\mgf{p}{\vX_i}
=%
\sum_{\bs \pi \in \Omega_i} p(\bs \pi) e^{\bs t^T \vX_i(\bs \pi)},
\end{equation}
for all $\bs t \in \mb R^D$ and the cross-moment can be retrieved
from the \textit{MGF} by differentiating:
\begin{equation}
\label{mgf: m_nu_sum}
\mom{\nu}{i}=%
\frac{\pd^{|\bs \nu|} \mgf{p}{\vX_i}}{\pd^{\nu_1}t_1 \dots
\pd^{\nu_D}t_d} \big |_{\bs t = \bs 0}= \mc D_{\bs \nu}\Big\{
\mgf{p}{\vX_i} \Big\},
\end{equation}
The \emph{$i$-th conditional cross-moment of an order $\bs \nu$},
$\momc{\nu}{i}{u}$, and the \textit{$i$-th conditional moment
generating function}, $\mgfc{p}{\bs X}{\bs u}$ are defined in the
similar manner if the summing is performed over the set of all
derivations starting at $A_i$ and ending with a string $\bs u \in
\Sigma^*$ (note that in this case we are dealing with the finite
sums).

For the subgrammar $G_i =\big(\Sigma , \mN_i' , A_1 , \mR_i', p
\big)$, we can construct $\widetilde G_i =\big(\Sigma ,\mN_i' ,A_i
, \mR_i', w \big)$, the \textit{$i$-th moment-generating grammar},
with the weight function taking values from the semiring of
multivariate power series, $w : \mc R \ra \sR\big[ \sN_0^d \big]$,
defined with
\begin{equation}
\label{mgf: mu_tilde_rule} w (\pi)= p(\pi) e^{\bs t^T \bs Y(\pi)},
\end{equation}
for all $\pi \in \mc R$. A derivation $\bs \pi = \pi_1 \cdots
\pi_N$ in $G_i$ with the weight $p(\bs \pi)=p(\pi_1) \cdots
p(\pi_N)$ is also a derivation in $\widetilde G_i$, for which the
weight is given with
\begin{equation}
\label{mgf: mu_tilde_derivation}
w(\bs \pi)= w(\pi_1) \cdots w(\pi_N)=%
p(\pi_1) e^{\bs t^T \bs Y(\pi_1)}\cdots%
p(\pi_N) e^{\bs t^T \bs Y(\pi_N)}=%
p(\bs \pi) e^{\bs t^T \bs X(\bs \pi)}.
\end{equation}
The $i$-th \emph{MGF} can now be can be expressed as the sum of
derivation weights in $\widetilde G_i$ as
\begin{equation}
\label{mgf: M_X_sum}
\mgf{p}{\bs X_i}=%
\sum_{\bs \pi \in \Omega} p(\bs \pi) e^{\bs t^T \bs X_i(\bs \pi)}=%
\sum_{\bs \pi \in \Omega} w(\bs \pi).
\end{equation}
Thus, the problem of \textit{MGF} computation is reduced to the
problem of the \textit{partition function} computation
\cite{Nederhof_Satta_08a}, and the conditional \textit{MGF} can be
computed using the \textit{inside algorithm} \cite{Goodman_99}
over the binomial semiring \cite{Ilic_et_al_12}. In the following
sections we show how the expressions for the cross-moments and
conditional cross-moments can be derived from (\ref{mgf:
M_X_sum}).

\section{Cross-moments computation of \textit{SCFG}}
\label{cm}

Let $\widetilde G_i = \big(\Sigma, \mN_i', A_i, \mR_i' , w \big)$
be a weighted context-free grammar over a commutative semiring
$\big(\mb K, +, \cdot, 1, 0 \big)$ endowed with a topology $\tau$.
Assuming that for $1 \leq i \leq |\mN|$ the infinite collections
$\big\{ w(\bs \pi) \big\}_{\bs \pi \in \Omega_i}$ are summable in
$\tau$ and that the distributive law for infinite sums holds, we
define the \emph{partition function} $Z: \mN \ra \mb K$, which to
every nonterminal $A_i \in \mN$ associates the sum
\begin{equation}
\label{cm: Z_i_def}%
Z_i=\sum_{\bs \pi \in \Omega_i} w\big(\bs \pi\big).
\end{equation}
By factoring out the first rewriting of each derivation in the sum, using the distributive law, the partition function can be expressed with the system \cite{Nederhof_Satta_08a}:
\begin{equation}
\label{cm: Z_i_eq}
Z_i = \sum_{j=1}^{|\mR_i|}
w(A_i \ra B_{i,j})
\cdot \prod_{k=1}^{|\mN|} Z_k^{r_k(B_{i,j})}, 
\end{equation}
where $1 \leq i \leq |\mN|$.

Now, let  $\widetilde G_i = \big(\Sigma, \mN_i', A_i', \mR_i' , w
\big)$ be the moment-generating grammar for $G_i = \big(\Sigma,
\mN_i', A_i, \mR_i' , p \big)$ with
\begin{equation}
\label{cm: mu_tilde_rule} w (\pi)= p(\pi) e^{\bs t^T \bs Y(\pi)}
\end{equation}
and
\begin{equation}
\label{cm: X = sum Y}
\bs X_i(\pi_1 \cdots  \pi_N) =%
\bs Y\big( \pi_1 \big) + \cdots + \bs Y\big( \pi_N \big),
\end{equation}
for all $i=1, \dots, |\mc N|$. According to the discussion made in
section \ref{mgf}, the value of the partition function at the
nonterminal $A_i$ corresponds to the $i$-th moment-generating
function, $Z_i=\mgf{p}{\vX_i}$, and the $i$-th cross-moment
\begin{equation}
\label{cm: m_i^alpha}
\mom{\alpha}{i} =%
\mc D_{\bs \alpha}\big\{M_{\bs X}^{(i)}\big\}=%
\mc D_{\bs \alpha}\big\{Z_i\big\}=%
\sum_{\bs \pi \in \bs \Omega_i}p(\bs \pi) \bs X(\bs \pi)^{\bs
\alpha}
\end{equation}
can be computed by differentiating (\ref{cm: Z_i_eq}) and solving
the resulting equation. Note that
\begin{equation}
\label{cm: mom_0_i}
\mom{0}{i} =\mc D_{\bs 0}\big\{Z_i\big\} = %
\Big(\sum_{\bs \pi \in \Omega_i} p(\bs \pi) e^{\bs t^T \bs X(\bs \pi)}\Big)|_{\bs t = \bs 0}=%
\sum_{\bs \pi \in \Omega_i} p(\bs \pi) = 1,
\end{equation}
for all $1\leq i \leq |\mN|$.~The cross-moments of higher order
can be obtained by applying the generalized Leibniz's formula
(\ref{prel: glr: D_FG}) to (\ref{cm: Z_i_eq}), which leads us to
the following system:
\begin{equation}
\label{cm: mom_alpha_i_glr}
\mom{\alpha}{i} =
\sum_{j=1}^{|\mR_i|}
\sum_{\bs \beta \leq \bs \alpha} \bc{\bs \alpha}{\bs \beta}%
\mc D_{\bs \alpha - \bs \beta} \big\{ w\big(A_i \ra
B_{i,j}\big)\big\} \cdot \mc D_{\bs \beta}\Big\{
\prod_{k=1}^{|\mN|} Z_k^{r_k(B_{i,j})} \Big\},
\end{equation}
where
\begin{equation}
\label{cm: D_mu}
\mc D_{\bs \alpha - \bs \beta} \big\{ w\big(A_i \ra B_{i,j}\big)\big\}=%
p \big(A_i \ra B_{i,j}\big) \cdot \bs Y \big(A_i \ra B_{i,j}\big) ^{\bs \alpha - \bs \beta},
\end{equation}
since $w(\pi)=p(\pi)e^{\bs t^T \cdot \bs Y}$, for $\pi \in \mR$.
According to the generalized Leibniz's rule (\ref{prel: glr:
D_prod_F}), we have
\begin{equation}
\label{cm: D_beta_prod_Z_rk}
\mc D_{\bs \beta} \Big\{ \prod_{k=1}^{|\mN|} Z_k^{r_k(B_{i,j})} \Big\}= %
\sum_{\bs \gamma_1 +  \cdots + \bs \gamma_{|\mN| =\bs \beta}}%
\bc{\bs \beta}{\bs \gamma_1, \dots, \bs \gamma_{|\mN|}}%
\prod_{k=1}^{|\mN|}\mc D_{\bs \gamma_k}
\big\{ Z_k^{r_k(B_{i,j})} \big\}
\end{equation}
and
\begin{equation}
\label{cm: D_gamma_Z_rk}
\mc D_{\bs \gamma_k} \big\{ Z_k^{r_k(B_{i,j})} \big\} =
\mc D_{\bs \gamma_k} \Big\{ \prod_{l=1}^{r_k(B_{i,j})} Z_k \Big\} =
\sum_{\bs \delta_1 +  \cdots + \bs \delta_{r_k(B_{i,j})}=\bs \gamma_k}%
\bc{\bs \gamma_k}{\bs \delta_1, \dots, \bs \delta_{r_k(B_{i,j})}}%
\prod_{l=1}^{^{r_k(B_{i,j})}}%
\mom{{\bs \delta_l}}{k}.
\end{equation}
By substituting (\ref{cm: D_gamma_Z_rk}) and (\ref{cm:
D_beta_prod_Z_rk}) in (\ref{cm: mom_alpha_i_glr}), 
we obtain:
\begin{equation}
\label{cm: mom_alpha_i_sum_Q}
\mom{\alpha}{i}=
\sum_{j=1}^{|\mR_i|}\sum_{\bs \beta \leq \bs \alpha}
Q_{i,j}\big(\bs \alpha, \bs \beta\big),
\end{equation}
where
\begin{multline}
\label{cm: Q_alpha_beta}
Q_{i,j}\big(\bs \alpha,\bs \beta\big) =
\bc{\bs \alpha}{\bs \beta}
p \big(A_i \ra B_{i,j}\big) \cdot \bs Y \big(A_i \ra B_{i,j}\big) ^{\bs \alpha - \bs \beta} \cdot \\
\sum_{\bs \gamma_1 +  \cdots + \bs \gamma_{|\mN| =\bs \beta}}
\bc{\bs \beta}{\bs \gamma_1, \dots, \bs \gamma_{|\mN|}}  %
\prod_{k=1}^{|\mN|} 
\sum_{\bs \delta_1 +  \cdots + \bs \delta_{r_k(B_{i,j})}=\bs \gamma_k}%
\bc{\bs \gamma_k}{\bs \delta_1, \dots, \bs \delta_{r_k(B_{i,j})}}%
\prod_{l=1}^{^{r_k(B_{i,j})}} \mom{\delta_l}{k}.
\end{multline}
To solve the system (\ref{cm: mom_alpha_i_sum_Q}), we split it
into two parts: one depending and the other not depending  on
$\mom{\alpha}{i}$:
\begin{equation}
\label{cm: mom_split_sum_Q} \mom{\alpha}{i}=
\sum_{j=1}^{|\mR_i|}Q_{i,j}\big(\bs \alpha, \bs \alpha \big)+
\sum_{j=1}^{|\mR_i|}\sum_{\bs \beta < \bs \alpha} Q_{i,j}\big(\bs
\alpha, \bs \beta \big),
\end{equation}
where
\begin{equation}
\label{cm: Q_alpha_alpha}
Q_{i,j}\big(\bs \alpha, \bs \alpha \big)
= p \big(A_i\ra B_{i,j}\big) \cdot W_{i,j} \big(\bs \alpha\big)
\end{equation}
and
\begin{equation}
\label{cm: W_alpha_def}
W_{i,j}\big(\bs \alpha\big)=
\sum_{\gamma_1 +  \cdots + \gamma_{|\mN| =\bs \alpha}}%
\bc{\bs \alpha}{\bs \gamma_1, \dots, \bs \gamma_{|\mN|}} %
\prod_{k=1}^{|\mN|} %
\sum_{\bs \delta_1 +  \cdots + \bs \delta_{r_k(B_{i,j})}=\bs \gamma_k}%
\bc{\bs \gamma_k}{\bs \delta_1, \dots, \bs \delta_{r_k(B_{i,j})}}%
\prod_{l=1}^{^{r_k(B_{i,j})}} \mom{\delta_l}{k}.
\end{equation}
Further, if we set
\begin{equation}
\label{cm: H_ij_def}
H_{i,j}\big(\bs \gamma_1, \dots , \bs \gamma_{|\mN|}\big) =%
\bc{\bs \alpha}{\bs \gamma_1, \dots, \bs \gamma_{|\mN|}} %
\prod_{k=1}^{|\mN|} 
\sum_{\bs \delta_1 +  \cdots + \bs \delta_{r_k(B_{i,j})}=\bs \gamma_k}%
\bc{\bs \gamma_k}{\bs \delta_1, \dots, \bs \delta_{r_k(B_{i,j})}}%
\prod_{l=1}^{^{r_k(B_{i,j})}} %
\mom{\delta_l}{k},
\end{equation}
the expression for $W_{\bs \alpha}\big(B_{i,j}\big)$ can be rewritten as:
\begin{equation}
\label{cm: W_alpha_sum_H}
W_{i,j}\big(\bs \alpha\big)=
\sum_{n=1}^{|\mN|}%
H_{i,j}^{(n)}\big(\bs \gamma_1, \dots , \bs\gamma_{|\mN|}\big) +
\sum_{\substack{\bs \gamma_1 +  \cdots + \bs \gamma_{|\mN|} =\bs \alpha \\ \bs \gamma_1 ,  \dots , \bs \gamma_{|\mN|} < \bs \alpha}}%
H_{i,j}\big(\bs \gamma_1, \dots , \bs \gamma_{|\mN|}\big),
\end{equation}
where $H_{i,j}^{(n) } \big(\bs \gamma_1, \dots , \bs
\gamma_{|\mN|} \big)$ stands for $H_{i,j}\big(\bs \gamma_1, \dots
, \bs \gamma_{|\mN|}\big)$ with $\bs \gamma_n = \bs  \alpha$ and
all other $\bs \gamma$-s equal zero, which is, according to
(\ref{cm: H_ij_def}),
\begin{equation}
H_{i,j}^{(n)}\big(\bs \gamma_1, \dots , \bs \gamma_{|\mN|}\big) =%
\sum_{\bs \delta_1 +  \cdots + \bs \delta_{r_n(B_{i,j})}=\bs \alpha}%
\bc{\bs \alpha}{\bs \delta_1, \dots, \bs \delta_{r_n(B_{i,j})}}%
\prod_{l=1}^{^{r_n(B_{i,j})}} %
\mom{\delta_l}{n}\ \cdot
\prod_{\substack{k=1 \\ k \neq n}}^{|\mN|} 
\prod_{l=1}^{^{r_k(B_{i,j})}} \mom{0}{k}.
\end{equation}
Finally, after using of $\mom{0}{k} = 1$, we obtain
\begin{equation}
H_{i,j}^{(n)}\big(\bs \gamma_1, \dots , \bs \gamma_{|\mN|}\big) =%
\sum_{\bs \delta_1 +  \cdots + \bs \delta_{r_n(B_{i,j})}=\bs \alpha}%
\bc{\bs \alpha}{\bs \delta_1, \dots, \bs \delta_{r_n(B_{i,j})}}%
\prod_{l=1}^{^{r_n(B_{i,j})}} \mom{{\bs \delta_l}}{n},
\end{equation}
which can be rewritten using the same procedure as
\begin{multline}
\label{cm: H_ij_n_lin_mom}
H_{i,j}^{(n)}\big(\bs \gamma_1, \dots , \bs \gamma_{|\mN|}\big) =%
\sum_{s=1}^{r_n(B_{i,j})}%
\mom{{\bs \alpha}}{n}\ + \
\sum_{\substack{\bs \delta_1 +  \cdots + \bs \delta_{r_k(B_{i,j})} = \bs \alpha \\ %
\bs \delta_1,  \dots , \bs \delta_{r_n(B_{i,j})} < \bs \alpha}}
\bc{\bs \alpha}{\bs \delta_1, \dots, \bs \delta_{r_n(B_{i,j})}}%
\prod_{l=1}^{^{r_n(B_{i,j})}}
\mom{{\bs \delta_l}}{m} = \\
= r_n(B_{i,j}) \cdot
\mom{{\bs \alpha}}{n}\ + \
\sum_{\substack{\bs \delta_1 +  \cdots + \bs \delta_{r_n(B_{i,j})} = \bs \alpha \\ %
\bs \delta_1,  \dots , \bs \delta_{r_n(B_{i,j})} < \bs \alpha}}
\bc{\bs \alpha}{\bs \delta_1, \dots, \bs \delta_{r_n(B_{i,j})}}%
\prod_{l=1}^{^{r_n(B_{i,j})}}
\mom{{\bs \delta_l}}{n}.
\end{multline}
By substitution of (\ref{cm: H_ij_n_lin_mom}) in (\ref{cm:
W_alpha_sum_H}) it follows that:
\begin{multline}
\label{cm: W_alpha_lin _mom}
W_{i,j}\big(\bs \alpha\big)=
\sum_{n=1}^{|\mN|}%
r_n(B_{i,j}) \cdot
\mom{{\bs \alpha}}{n}\ + \\
\sum_{n=1}^{|\mN|}%
\sum_{\substack{\bs \delta_1 +  \cdots + \bs \delta_{r_n(B_{i,j})} = \bs \alpha \\ %
\bs \delta_1,  \dots , \bs \delta_{r_n(B_{i,j})} < \bs \alpha}}
\bc{\bs \alpha}{\bs \delta_1, \dots, \bs \delta_{r_n(B_{i,j})}}%
\prod_{l=1}^{^{r_n(B_{i,j})}}
\mom{{\bs \delta_l}}{n}\ +                                             
\sum_{\substack{\bs \gamma_1 +  \cdots + \bs \gamma_{|\mN|} =\bs \alpha \\ \bs \gamma_1 ,  \dots , \bs \gamma_{|\mN|} < \bs \alpha}}%
H_{i,j}\big(\bs \gamma_1, \dots , \bs\gamma_{|\mN|}\big).
\end{multline}
Further, by substitution of (\ref{cm: W_alpha_lin _mom}) and
(\ref{cm: Q_alpha_alpha}) in (\ref{cm: mom_split_sum_Q}), the
moment can be expressed with:
\begin{multline}
\label{cm: mom_lin_mom}
\mom{{\bs \alpha}}{i} =
\sum_{j=1}^{|\mR_i|} p (A_i \ra B_{i,j})
\sum_{n=1}^{|\mN|}%
r_n(B_{i,j}) \cdot
\mom{{\bs \alpha}}{n}\ + \\
\sum_{j=1}^{|\mR_i|} p (A_i \ra B_{i,j})
\sum_{n=1}^{|\mN|}%
\sum_{\substack{\bs \delta_1 +  \cdots + \bs \delta_{r_n(B_{i,j})} = \bs \alpha \\ %
\bs \delta_1,  \dots , \bs \delta_{r_n(B_{i,j})} < \bs \alpha}}
\bc{\bs \alpha}{\bs \delta_1, \dots, \bs \delta_{r_n(B_{i,j})}}%
\prod_{l=1}^{^{r_n(B_{i,j})}}
\mom{{\bs \delta_l}}{n}\ + \\                                            
\sum_{j=1}^{|\mR_i|} p (A_i \ra B_{i,j})
\sum_{\substack{\bs \gamma_1 +  \cdots + \bs \gamma_{|\mN|} =\bs \alpha \\ \bs \gamma_1 ,  \dots , \bs \gamma_{|\mN|} < \bs \alpha}}%
H_{i,j}\big(\bs \gamma_1, \dots , \bs\gamma_{|\mN|}\big) +
\sum_{j=1}^{|\mR_i|}\sum_{\bs \beta < \bs \alpha}%
Q_{i,j}\big(\bs \alpha , \bs \beta\big),
\end{multline}
where $H_{i,j}\big(\bs \gamma_1, \dots , \bs\gamma_{|\mN|}\big)$ and $Q_{i,j}\big(\bs \alpha, \bs \beta\big)$ are given
with (\ref{cm: H_ij_def}) and (\ref{cm: Q_alpha_beta}). Finally, if we introduce
\begin{multline}
\label{cm: c_alpha_i}
c_{i}^{(\bs \alpha)}=
\sum_{j=1}^{|\mR_i|} p (A_i \ra B_{i,j})
\sum_{n=1}^{|\mN|}%
\sum_{\substack{\bs \delta_1 +  \cdots + \bs \delta_{r_n(B_{i,j})} = \bs \alpha \\ %
\bs \delta_1,  \dots , \bs \delta_{r_n(B_{i,j})} < \bs \alpha}}
\bc{\bs \alpha}{\bs \delta_1, \dots, \bs \delta_{r_n(B_{i,j})}}%
\prod_{l=1}^{^{r_n(B_{i,j})}}
\mom{{\bs \delta_l}}{n}\ + \\                                            
\sum_{j=1}^{|\mR_i|} p (A_i \ra B_{i,j})
\sum_{\substack{\bs \gamma_1 +  \cdots + \bs \gamma_{|\mN|} =\bs \alpha \\ \bs \gamma_1 ,  \dots , \bs \gamma_{|\mN|} < \bs \alpha}}%
H_{i,j}\big(\bs \gamma_1, \dots , \bs\gamma_{|\mN|}\big) +
\sum_{j=1}^{|\mR_i|}\sum_{\bs \beta < \bs \alpha}
Q_{i,j}\big(\bs \alpha, \bs \beta\big),
\end{multline}
the equation (\ref{cm: mom_lin_mom}) can be more compactly written as:
\begin{equation}
\mom{{\bs \alpha}}{i} = \sum_{n=1}^{|\mN|} \text M_{i,n} \cdot \mom{{\bs \alpha}}{n} + c_{i}^{(\bs \alpha)},
\end{equation}
or in a matrix form:
\begin{equation}
\momvec{p}{\mia} =\text M \cdot \momvec{p}{\mia} + \bs c^{(\mia)},
\end{equation}
where $\momvec{p}{\mia} = \big[\mom{\alpha}{1}, \dots ,
\mom{\alpha}{|\mN|}\big]^T$ is the \textit{cross-moment vector}
$\bs c^{(\bs \alpha)}= \big[c^{(\bs \alpha)}_1, \dots, c^{(\bs
\alpha)}_{|\mN|} \big]^T$ and $\text M$ is the momentum matrix
defined in Theorem \ref{cm: theo: consistency}.~Since we assume
that the condition $\rho(\text M) < 1$ given in Theorem \ref{cm:
theo: consistency} is satisfied, $\text I - \text M$ is
invertible, and the matrix equation has a unique solution given
with
\begin{equation}
\label{cm: moment system} \momvec{p}{\mia} = \big(\text I - \text
M \big)^{-1} \bs c^{(\bs \alpha)} .
\end{equation}
Provided that the we have computed the inverse $\big(\text I -
\text M \big)^{-1}$, which does not depend on $\bs \alpha$, the
cross-moment is completely determined by the term $\bs c^{(\bs
\alpha)}$, which depends on all cross-moments of the order lower
than $\bs \alpha$ and can be computed using (\ref{cm: c_alpha_i}).
In the following section, we derive $\bs c^{(\bs \alpha)}$ for
scalar random variables up to the second order, and retrieve the
previous results for the first and second order moments
\cite{Booth_Thompson_73}, \cite{Hutchins_72} as a special case of
the equation (\ref{cm: moment system}).

\subsection{First order moments}

In the case of the first order moments $\bs \alpha = (1)$
and the expression (\ref{cm: m_i^alpha}) reduces to the
expectation of $\bs X_i$,
\begin{equation}
\label{cm: 1: m_i^1} \mom{1}{i} = \sum_{\bs \pi \in \Omega_i}
p(\bs \pi) \bs X_i(\bs \pi).
\end{equation}
The moment vector, $\momvec{p}{\mia} = \big[\mom{1}{1}, \dots ,
\mom{1}{|\mc N|}\big]$, is computed as in the equation (\ref{cm:
moment system}),
\begin{equation}
\label{cm: 1: moment system} \momvec{p}{1} = \big(\text I - \text
M \big)^{-1} \bs c^{(1)},
\end{equation}
where  $c^{(1)}= \big[c^{(1)}_1, \dots, c^{(1)}_{|\mN|} \big]^T$.
The first and second sum in the expression (\ref{cm: c_alpha_i})
for $c^{(\bs \alpha)}_{i}$ reduce to zero and $c_{i}^{(1)}=
\sum_{j=1}^{|\mR_i|} Q_{i,j}(1, 0)$, or, after the use of the
expression (\ref{cm: Q_alpha_beta}) for $Q_{i,j}(\bs \alpha, \bs
\beta)$,
%
\begin{equation}
\label{cm: m1: c_1_i}
c_{i}^{(1)}=
\sum_{j=1}^{|\mR_i|} %
p \big(A_i\ra B_{i,j}\big) \cdot \bs Y \big(A_i\ra B_{i,j}\big).
\end{equation}

Let, $\pi_1 \cdots \pi_N$ be a derivation starting at the start
symbol $A_1$ and ending with a string $\bs u \in \Sigma^*$. If we
set $\bs Y \big(A_i\ra B_{i,j}\big) = 1$, according to (\ref{cm: X
= sum Y}), we have $\bs X_1(\pi_1 \cdots \pi_N) = \sum_{n=1}^N \bs
Y\big( \pi_n \big) = N$, i.e. $\bs X_1$ is the length of the
derivation. According to the expression (\ref{cm: 1: m_i^1}), the
moment $\mom{1}{1}$ is the expected derivation length which agrees
with \cite{Booth_Thompson_73} and \cite{Hutchins_72}.

Similarly, if we set $\bs Y \big(A_i\ra B_{i,j}\big) =
\sum_{n=1}^{|\Sigma|} t_n (i,j)$, where $t_n (i,j)$ denotes the
number of terminals in the string $B_{i,j}$, the variable $\bs
X_1(\pi_1 \cdots \pi_N)$ reduces to the length of the word derived
from $\pi_1 \cdots \pi_N$. In this case, the moment $\moms{1}{1}$
reduces to the expected string length and the formula (\ref{cm:
m1: c_1_i}) reduces to the result from \cite{Booth_Thompson_73}.

\subsection{Second order moments}

The formula for the second order moments is somewhat more
complicated.~In the case when $\bs \alpha=(2)$, $c^{(\bs
\alpha)}_{i}$ is reduced to:
\begin{multline}
\label{cm: m2: c_def}
c^{(2)}_{i}=
\sum_{j=1}^{|\mR_i|} p (A_i \ra B_{i,j})
\sum_{n=1}^{|\mN|}%
\sum_{\substack{\bs \delta_1 +  \cdots + \bs \delta_{r_n(B_{i,j})} = \bs \alpha \\ %
\bs \delta_1,  \dots , \bs \delta_{r_n(B_{i,j})} < 2}}
\bc{2}{\bs \delta_1, \dots, \bs \delta_{r_n(B_{i,j})}}%
\prod_{l=1}^{^{r_n(B_{i,j})}}
\mom{{\bs \delta_l}}{n}\ + \\                                            
\sum_{j=1}^{|\mR_i|} p (A_i \ra B_{i,j})
\sum_{\substack{\gamma_1 +  \cdots + \gamma_{|R|} =2 \\ \gamma_1 ,  \dots , \gamma_{|R| < 2}}}%
H_{i,j}\big(\bs \gamma_1, \dots , \bs\gamma_{|R|}\big) +
\sum_{j=1}^{|\mR_i|} Q_{i,j}\big(2, 0\big)+
\sum_{j=1}^{|\mR_i|}Q_{i,j}\big(2, 1\big).
\end{multline}
The first sum in the previous expression can be transformed to:
\begin{multline}
\label{cm: m2: first_sum}
\sum_{j=1}^{|\mR_i|} p (A_i \ra B_{i,j})
\sum_{n=1}^{|\mN|}%
\sum_{\substack{\bs \delta_1 +  \cdots + \bs \delta_{r_n(B_{i,j})} = 2 \\ %
\bs \delta_1,  \dots , \bs \delta_{r_n(B_{i,j})} < 2}}
\bc{2}{\bs \delta_1, \dots, \bs \delta_{r_n(B_{i,j})}}%
\prod_{l=1}^{{r_n(B_{i,j})}}
\mom{{\bs \delta_l}}{n} = \\
\sum_{j=1}^{|\mR_i|} p (A_i \ra B_{i,j})
\sum_{n=1}^{|\mN|}%
r_n(B_{i,j})\big(r_n(B_{i,j}) - 1 \big) \cdot
\big(\moms{1}{n}\big)^2.
\end{multline}
To compute the second sum we introduce $H_{i,j}^{(a,b)}\big(\bs
\gamma_1, \dots , \bs \gamma_{|\mN|}\big)$, which is
$H_{i,j}\big(\bs \gamma_1, \dots , \bs\gamma_{|\mN|}\big)$ with
$\bs \gamma_a=\bs \gamma_b = 1$ and with all other $\bs \gamma$-s
equals to zero. We have:
\begin{multline}
H_{i,j}^{(a,b)}\big(\bs \gamma_1, \dots , \bs \gamma_{|\mN|}\big) =%
2 \cdot
\sum_{\delta_1 +  \cdots + \delta_{r_a(B_{i,j})}=\bs \gamma_a}%
\bc{\bs \gamma_a}{\bs \delta_1, \dots, \bs \delta_{r_a(B_{i,j})}}%
\prod_{l=1}^{{r_a(B_{i,j})}}
\moms{{\bs \delta_l}}{k} \\
\sum_{\delta_1 +  \cdots + \delta_{r_b(B_{i,j})}=\bs \gamma_b}%
\bc{\bs \gamma_b}{\bs \delta_1, \dots, \bs \delta_{r_b(B_{i,j})}}%
\prod_{l=1}^{{r_b(B_{i,j})}}
\moms{{\bs \delta_l}}{a} \cdot 
\prod_{\substack{k=1 \\ k \neq a,b }}^{|\mN|} 
\sum_{\bs \delta_1 +  \cdots + \bs \delta_{r_k(B_{i,j})}=\bs \gamma_k}%
\bc{\bs \gamma_k}{\bs \delta_1, \dots, \bs \delta_{r_k(B_{i,j})}}%
\prod_{l=1}^{^{r_k(B_{i,j})}}
\mom{{\bs \delta_l}}{b},
\end{multline}
and
\begin{equation}
\label{cm: m2: H_ij_ab}
H_{i,j}^{(a,b)}\big(\bs \gamma_1, \dots , \bs \gamma_{|\mN|}\big) =%
2 \cdot \sum_{c=1}^{r_a(B_{i,j})} \moms{1}{k} \cdot \sum_{d=1}^{r_b(B_{i,j})} \moms{1}{k}=%
2 \cdot r_a(B_{i,j})\cdot r_b(B_{i,j}) \cdot \moms{1}{a} \moms{1}{b}.
\end{equation}
By substitution of the second sum in (\ref{cm: m2: c_def})
\begin{multline}
\label{cm: m2: sec_sum}
\sum_{j=1}^{|R_i|} p (A_i \ra B_{i,j})
\sum_{\substack{\gamma_1 +  \cdots + \gamma_{|\mN|} =2 \\ \gamma_1 ,  \dots , \gamma_{|\mN| < 2}}}%
H_{i,j}\big(\bs \gamma_1, \dots , \bs\gamma_{|\mN|}\big)=
\sum_{j=1}^{|\mR_i|} p (A_i \ra B_{i,j})
\sum_{a=1}^{|\mN|}\sum_{b=a+1}^{|\mN|}
H_{i,j}^{(a,b)}\big(\bs \gamma_1, \dots , \bs\gamma_{|\mN|}\big)= \\
=2 \cdot \sum_{j=1}^{|\mR_i|} p (A_i \ra B_{i,j})
\sum_{a=1}^{|\mN|}\sum_{b=a+1}^{|\mN|}
r_a(B_{i,j})\cdot r_b(B_{i,j}) \cdot \moms{1}{a} \moms{1}{b}=\\
=\sum_{j=1}^{|\mR_i|} p (A_i \ra B_{i,j})
\sum_{a=1}^{|\mN|}\sum_{b=1}^{|\mN|}
r_a(B_{i,j})\cdot r_b(B_{i,j}) \cdot \moms{1}{a} \moms{1}{b}- 
\sum_{j=1}^{|\mR_i|} p (A_i \ra B_{i,j}) \sum_{n=1}^{|\mN|}
r_n(B_{i,j})^2 \big(\moms{1}{n}\big)^2.
\end{multline}
Now, (\ref{cm: m2: c_def}) reduces to
\begin{equation}
\label{cm: m2: c_CR_Q} c^{(2)}_i=CR_i+ \sum_{j=1}^{|\mR_i|}
Q_{i,j}\big(2, 0\big)+ \sum_{j=1}^{|\mR_i|}Q_{i,j}\big(2, 1),
\end{equation}
where
\begin{multline}
\label{cm: m2: CR_i}
CR_i=
\sum_{j=1}^{|\mR_i|} p (A_i \ra B_{i,j})
\sum_{a=1}^{|\mN|}\sum_{b=1}^{|\mN|}
r_a(B_{i,j})\cdot r_b(B_{i,j}) \cdot \moms{1}{a} \moms{1}{b}- 
\sum_{j=1}^{|\mR_i|} p (A_i \ra B_{i,j}) \sum_{n=1}^{|\mN|}
r_n(B_{i,j}) \big(\moms{1}{n}\big)^2
\end{multline}
and
\begin{equation}
\label{cm: m2: Q_20}
Q_{i,j}\big(2, 0) =
p \big(A_i\ra B_{i,j}\big) \cdot %
Y \big(A_i\ra B_{i,j}\big)^2,
\end{equation}
\begin{equation}
\label{cm: m2: Q_21}
Q_{i,j}\big(2, 1) =
2 \cdot p \big(A_i\ra B_{i,j}\big) \cdot %
Y \big(A_i\ra B_{i,j}\big)
\sum_{n=1}^{|\mN|} \sum_{a=1}^{r_n(B_{i,j})}
\moms{1}{n}=
2 \cdot p \big(A_i\ra B_{i,j}\big) \cdot %
Y \big(A_i\ra B_{i,j}\big)
\sum_{n=1}^{|\mN|} r_n(B_{i,j})
\moms{1}{n}.
\end{equation}

If we set $Y \big(A_i\ra B_{i,j}\big)=1$ for all $A_i\ra B_{i,j}
\in \mR$, $\bs X_1$ becomes derivation length.~The formula for
computing the second order moments of derivation length is given
in \cite{Hutchins_72} and it can be derived from the equation
(\ref{cm: m2: c_CR_Q}), since
\begin{align}
\label{cm: m2: sum_Q_20}
&\sum_{j=1}^{|\mR_i|} Q_{i,j}\big(2, 0) =
\sum_{j=1}^{|\mR_i|}  p \big(A_i\ra B_{i,j}\big) \cdot %
Y \big(A_i\ra B_{i,j}\big)^2 =
\sum_{j=1}^{|\mR_i|}  p \big(A_i\ra B_{i,j}\big)=
1,\\
\label{cm: m2: sum_Q_21}
&\sum_{j=1}^{|\mR_i|}
Q_{i,j}\big(2, 1) =
2 \cdot \sum_{n=1}^{|\mN|}
\Big(\sum_{j=1}^{|\mR_i|}
p \big(A_i\ra B_{i,j}\big) \cdot
r_n(B_{i,j})\Big) \moms{1}{n}=
2 \cdot \sum_{n=1}^{|\mN|}
e_{i,n} \moms{1}{n}=
2\cdot \moms{1}{n} - 2,
\end{align}
where the last equation follows from (\ref{cm: moment system}), and
\begin{equation}
\label{cm: m2: c_final}
c^{(2)}_i=CR_i + 2\cdot \moms{1}{n} - 1.
\end{equation}
Finally, by substituting (\ref{cm: m2: c_final}) in (\ref{cm:
moment system}), we obtain
\begin{equation}
\text m^{(\bs \alpha)}\big\{\bs X\big\} = \big(\text I - \text M \big)^{-1} \cdot
\Big(\text{CR}_i + 2 \cdot \text m_1 - \textbf{1} \Big),
\end{equation}
where $\text{CR}_i = \big[CR_1, \dots , CR_{|\mN|}\big]$ and $\textbf{1} = \big[ 1, \dots , 1 \big]$, in agreement with \cite{Hutchins_72}.

\section{Conditional cross-moments computation for SCFGs}

Let $\widetilde G_i = \big(\Sigma, \mN_i', A_i, \mR_i' , w \big)$
be a weighted context-free grammar over a commutative semiring
$\big(\mb K, +, \cdot, 1, 0 \big)$, and $\Omega_i(\bs u)$ be a set
of all derivations which derive $\bs u \in \Sigma^*$ starting from
$A_i$.~The \textit{inside weight} of the weighted grammar
$\widetilde G_i$ is the function $\sigma_i: \mc N_i \times
\Sigma^* \ra \sR\big[ \sN_0^d \big]$, defined as the sum of
weights of all derivations starting with $A_i$ and ending with
$\bs u$,
\begin{equation}
\label{ccm: sigma_def}%
\sigma_i \big(\bs u \big) = \sum_{\bs \pi \in \Omega_i(\bs u)} w
\big(\bs \pi \big),
\end{equation}
for $1 \leq i \leq |\mc R|$ and $\bs u \in \Sigma^*$. Let $A_i \ra
B_{i,j} \in \mc R$ and
\begin{equation}
\label{cmc: B_ij_def} B_{i,j} = \bs v_1 A_{i_1} \bs v_2 A_{i_2}
\cdots \bs v_k A_{i_k} v_{k+1},
\end{equation}
where $\bs v_i \in \Sigma^*$ and $A_{i_n} \in \mc N$. 
For the cycle-free reduced grammars the inside weight can be
computed using the \emph{inside algorithm} \cite{Goodman_99} and
\cite{Tendeau_98} which, after recursive application of
\begin{equation}
\label{cmc: sigma_rec}
\sigma_i(\bs u) = \sum_{j=1}^{|\mc R_i|} 
\sum_{ \substack{\bs u_1 ,\bs u_2 , \dots, \bs u_k \in \Sigma^* \\ \bs u = \bs v_1 \bs u_1 \bs v_2 \cdots \bs v_k \bs u_k \bs v_{k+1}}}%
w(A_i \ra B_{i,j}) \cdot \prod_{j=1}^k \sigma_{i_j}(\bs u_j),
\end{equation}
ends with the equation in which only rules $A_i \ra \bs u, \bs u
\in \Sigma^*$ appear on the right-hand side:
\begin{equation}
\label{cmc: sigma_base}%
\sigma_i \big(\bs u \big) = w \big( A_i \ra \bs u \big).
\end{equation}
Note that in practice, the recursive algorithm defined by the
equations (\ref{cmc: sigma_base})-(\ref{cmc: sigma_base}) is
always implemented in the iterative manner using some of the
parsing techniques considered in \cite{Goodman_99}.

The goal of this section is an efficient computation of $i$-th
conditional cross-moments of an order $\bs
\nu=(\nu_1,\dots,\nu_D)$,
\begin{equation}
\label{cond mgf: m_nu_def} \momc{\nu}{i}{\bs u}=
\sum_{\bs \pi \in \Omega_i(\bs u)} p(\bs \pi) \cdot %
X_{i,1}(\bs \pi)^{\nu_1} \cdots X_{i,D}(\bs \pi)^{\nu_d} = %
\sum_{\bs \pi \in \Omega_i(\bs u)} p(\bs \pi)\ \bs X_i(\bs
\pi)^{\bs \nu},
\end{equation}
using the inside algorithm. Let $\widetilde G_i =\big(\Sigma ,
\mN_i' , A_i , \mR_i', w \big)$ be the \emph{$i$-th moment
generating grammar} for $G_i =\big(\Sigma , \mN_i' , A_i , \mR_i',
p \big)$, where the weight takes values from the semiring of
formal power series $w : \mc R \ra \sR \big[\sN_0^d \big]$,
defined with
\begin{equation}
w (\pi)= p(\pi) e^{\bs t^T \bs Y(\pi)},
\end{equation}
for all $\pi \in \mc R$.
Then, the conditional $MGF$ $\mgfc{p}{\vX_i}{\bs u}$ is the
\emph{inside} weight in the semiring of power series
\begin{equation}
\sigma_i \big(\bs u \big)=%
\egf{\mia}{\mu_{p, \vX_{i} | \bs u}},
\end{equation}
and it can be expressed with a recursive equation given with the
expressions (\ref{cmc: sigma_rec}) and (\ref{cmc: sigma_base}). On
the other hand,
\begin{equation}
\mgfc{p}{\vX_i}{\bs u}=%
\egf{\mia}{\mu_{p, \vX_{i} | \bs u}},
\end{equation}
and the conditional cross-moments can be obtained by applying the
map $\mc B^{(\minu)}: \sR(\vt) \ra \sR^{|\Anu|}$ on $\sigma_i
\big(\bs u \big)$, which is defined with
\begin{equation}
\hbegf{\minu}{\mia}{z}= \bsel{z}{\mia}{\Anu},
\end{equation}
and we have
\begin{equation}
\mc B^{(\minu)}%
\big\{ \sigma_i \big(\bs u \big) \big\}=%
\bsel{\mu_{p, \vX_{i} | \bs u}}{\mia}{\Anu}.
\end{equation}
The mapping $\mc B^{(\minu)}$ maps the semiring of the power
series in the binomial semiring of an order $\bs \nu$, which is
defined as the tuple $(\sR^{|\cA_{\bs \nu}|}, \oplus, \ot, \bs 0,
\bs 1)$, where the $\oplus$ and $\otimes$ are defined with
\begin{align}
&u \oplus v = \Big(u^{(\mia)} +  v^{(\mia)} \Big)_{\mia \in A_{\bs \nu}}, \\
u \otimes v &= \Big(\ \sum_{\mib \leq \mia} \bc{\mia}{\mib} \
u^{(\mib)} \cdot v^{(\mia - \mib)}\ \Big)_{\mia \in A_{\bs \nu}},
\end{align}
for all $u,v \in \sR^{|\cA_{\bs \nu}|}$, and the identities for
$\oplus$ and $\otimes$ are respectively given with
\begin{align*}
\bs 0 &=(\ \underbrace{ 0, 0, \dots , 0}_{|\sA_{\bs \nu}| \text{ times}}\ )\\
\bs 1 &=(1, \underbrace{ 0, \dots , 0}_{|\sA_{\bs \nu}| - 1 \text{
times}}).
\end{align*}
In this way, the cross-moments of the order lower than or equal to
$\minu$ can recursively be expressed as
\begin{equation}
\label{cmc: cond sigma_rec}
\mc B^{(\minu)}\big\{\sigma_i(\bs u)\big\} = %
\osum_{j=1}^{|\mc R_i|} 
\osum_{ \substack{\bs u_1 ,\bs u_2 , \dots, \bs u_k \in \Sigma^* \\ \bs u = \bs v_1 \bs u_1 \bs v_2 \cdots \bs v_k \bs u_k \bs v_{k+1}}}%
w(A_i \ra B_{i,j}) \ot \oprod_{j=1}^k%
\mc B^{(\minu)}%
\big\{\sigma_{i_j}(\bs u_j)\big\}
\end{equation}
with the base case
\begin{equation}
\mc B^{(\minu)}\big\{\sigma_i \big(\bs u \big)\big\} =\mc
B^{(\minu)}\big\{ w \big( A_i \ra \bs u \big)\big\}.
\end{equation}
As shown in \cite{Ilic_et_al_12}, the following equalities hold:
\begin{align}
\label{pcfg: bsr: lemma addition}
&\Big( \osum_{n=1}^N w_n \Big)^{(\mia)}=%
\sum_{n=1}^N w^{(\mia)}_n,\\
\label{pcfg: bsr: lemma multipl}
\big( \oprod_{n=1}^N w_n \big)^{(\mia)}&=%
\sum_{\mib_1 + \cdots + \mib_N  = \mia} \bc{\mia}{\mib_1, \dots , \mib_N}%
\prod_{n=1}^N w^{(\mib_n)}_n,
\end{align}
for $w_n \in \sR^{|\cA_{\bs \nu}|};\ n=1, \dots, N$, and we get 
\begin{equation}
\label{cmc: mom_rec}
\momc{{\bs \alpha}}{i}{\bs u}= 
\sum_{j=1}^{|\mN|}\ %
\sum_{ \substack{\bs u_1 ,\bs u_2 , \dots, \bs u_k \in \Sigma \\ \bs u = \bs v_1 \bs u_1 \bs v_2 \cdots \bs u_k \bs v_k \bs u_{k+1}}}%
\sum_{ \beta \leq \alpha}%
\ \bc{\alpha}{\beta}\ %
p \big(A_i \ra B_{i,j}\big) \cdot %
\bs Y \big(A_i \ra B_{i,j}\big)^{\bs \alpha - \bs \beta}
\sum_{\bs \gamma_1 + \cdots \gamma_k = \beta} \prod_{j=1}^k%
\bc{\bs \beta}{\bs \gamma_1, \dots, \bs \gamma_{k}} \
\momc{{\gamma_j}}{i_j}{\bs u_j}
\end{equation}
with the base case
\begin{equation}
\label{cmc: mom_base}%
\momc{{\bs \gamma}}{i}{{\bs u}}=%
p \big(A_i \ra \bs u \big) \cdot \bs Y \big(A_i \ra \bs u \big)
^{\bs \gamma}.
\end{equation}
As previously mentioned, the recursive algorithm (\ref{cmc:
mom_rec})-(\ref{cmc: mom_base}) can always be implemented in the
iterative manner using some of the procedures considered in
\cite{Goodman_99}.

The algorithm given by equations (\ref{cmc: mom_rec})-(\ref{cmc:
mom_base}) can be considered as a generalization of
the algorithms by Li and Eisner \cite{Li_Eisner_09} for the
cross-moments of order $\bs \alpha = (1, 1)$ and Hwa \cite{Hwa_00}
for the cross-moments of order $\bs \alpha = (1)$.~Li and Eisner
introduced the second order entropy semiring, which is the
binomial semiring of the order $(1,1)$, and ran the inside
algorithm on it. 
The algorithm for the moments of order $\bs \alpha=(1)$ is
provided by Hwa \cite{Hwa_00}, where con\-ditional entropy is
considered.~As noted in \cite{Cohen_11}, Hwa's algorithm can be
obtained by running the inside algorithm over the first order
entropy semiring \cite{Cortes_et_al_08}, which is the binomial
semiring of the order $(1)$. Hwa's algorithm is considered in the
following subsection.

\subsection{First order conditional moments}
\smallskip

In the case of first order conditional moments $\bs \alpha=(1)$,
the $i$-th conditional cross-moment (\ref{cond mgf: m_nu_def}) is
the expectation of $\bs X_i$,
\begin{equation}
\momcs{1}{1}{{\bs u}}=%
\sum_{\bs \pi \in \Omega_i(\bs u)} p(\bs \pi)\bs X_i(\bs \pi).
\end{equation}
In this case, the recursive equations
(\ref{cmc: mom_rec})-(\ref{cmc: mom_base}) reduce to
\begin{equation}
\label{cmc: 1: mom_rec}
\momcs{0}{i}{\bs u}= 
\sum_{j=1}^{|\mN|}\ %
\sum_{ \substack{\bs u_1 ,\bs u_2 , \dots, \bs u_k \in \Sigma \\ \bs u = \bs v_1 \bs u_1 \bs v_2 \cdots \bs u_k \bs v_k \bs u_{k+1}}}%
p \big(A_i \ra B_{i,j}\big) \cdot %
\prod_{j=1}^k \momcs{0}{i_j}{\bs u_j}, %
\end{equation}
\begin{multline}
\label{cmc: 1: mom_rec}
\momcs{{1}}{i}{\bs u}= 
\sum_{j=1}^{|\mN|}\ %
\sum_{ \substack{\bs u_1 ,\bs u_2 , \dots, \bs u_k \in \Sigma \\ \bs u = \bs v_1 \bs u_1 \bs v_2 \cdots \bs u_k \bs v_k \bs u_{k+1}}}%
p \big(A_i \ra B_{i,j}\big) \cdot \bs Y \big(A_i \ra B_{i,j}\big)
\cdot\prod_{j=1}^k \momcs{0}{i_j}{\bs u_j}+ \\ %
p \big(A_i \ra B_{i,j}\big) \cdot \sum_{n=1}^{k}
\momcs{1}{i_n}{\bs u_n}%
\prod_{\substack{j=1 \\ j\neq n}}^k \momcs{0}{i_j}{\bs u_j},
\end{multline}
with the base case:
\begin{equation}
\momcs{0}{i}{{\bs u}}=%
p \big(A_i \ra \bs u \big),\quad\quad\quad
\momcs{1}{i}{{\bs u}}=%
p \big(A_i \ra \bs u \big) \cdot Y \big(A_i \ra \bs u \big).
\end{equation}

In \cite{Hwa_00}, Hwa considered the conditional entropy of the
grammar given in Chomsky form for which $B_{i,j}=v_1 A_{i_1} v_2
A_{i_2} v_3$ and $v_1,v_2,v_3$ are equal to the empty string.~The
conditional entropy is obtained as the moment $\momcs{1}{1}{{\bs
u}}$, where $\bs X_1(\bs \pi)= -\log p(\bs \pi)$, for all $\bs \pi
\in \Omega_1$, and Hwa's algorithm can be retrieved by imposing
Chomsky form condition in (\ref{cmc: mom_rec})-(\ref{cmc:
mom_base}), with $\bs Y(\pi_i)=- \log p(\pi_i)$.

\section{Conclusion}

In this paper we considered the problem of computing the
cross-moments and the conditional cross-moments of a vector
variable represented by a stochastic context-free grammar.~We
proposed new algorithms, which are derived by differentiation of
the recursive equations for the moment-generating function
\cite{Lund_04}, which are obtained from the algorithms for
computing the partition function of a \textit{SCFG}
\cite{Nederhof_Satta_08a} for the cross-moments and with the
inside algorithm \cite{Lari_Young_90}, \cite{Goodman_99} for the
conditional cross-moments. In this way, we obtained the algorithms
which can be considered as a generalization of the previously
developed formulas for
moments~\cite{Hutchins_72},~\cite{Wetherell_80} and conditional
cross-moments \cite{Hwa_00}, \cite{Li_Eisner_09}.
%

The computation of cross-moments may be demanding and often
infeasible. The proposed method for its solution via the
computation of moment-generating function turned out to be a very
elegant and powerful way. In the future, we hope that this idea
can successfully be reused in the theory of formal languages for
the computation of cross moments of string and tree automata
\cite{Calera_Carrasco_98}, \cite{Cortes_Mohri_04}.

\section*{References}
\bibliographystyle{plain}
\bibliography{SCFG}

\end{document}